# *CALLOC*: Curriculum Adversarial Learning for Secure and Robust Indoor Localization

Danish Gufran and Sudeep Pasricha
Department of Electrical and Computer Engineering,
Colorado State University Fort Collins, CO, USA
{Danish.Gufran, Sudeep}@colostate.edu

*Abstract*— Indoor localization has become increasingly vital for many applications from tracking assets to delivering personalized services. Yet, achieving pinpoint accuracy remains a challenge due to variations across indoor environments and devices used to assist with localization. Another emerging challenge is adversarial attacks on indoor localization systems that not only threaten service integrity but also reduce localization accuracy. To combat these challenges, we introduce *CALLOC*, a novel framework designed to resist adversarial attacks and variations across indoor environments and devices that reduce system accuracy and reliability. *CALLOC* employs a novel adaptive curriculum learning approach with a domain specific lightweight scaled-dot product attention neural network, tailored for adversarial and variation resilience in practical use cases with resource constrained mobile devices. Experimental evaluations demonstrate that *CALLOC* can achieve improvements of up to 6.03x in mean error and 4.6x in worst-case error against state-of-the-art indoor localization frameworks, across diverse building floorplans, mobile devices, and adversarial attacks scenarios.

## I. INTRODUCTION

Indoor localization involves the precise determination of the location of devices or individuals within indoor spaces. It has extensive applications across many domains, including asset management, assistive healthcare, navigation within buildings, and the provision of context-aware services [1]. Tech giants such as Apple, Google, and Microsoft are heavily investing in indoor localization research to enhance the accuracy and reliability of indoor location-based services [2]. However, achieving high-location accuracy in real-world indoor scenarios remains an open challenge [1].

Traditional navigation systems, such as Global Positioning System (GPS), heavily rely on satellite signals and clear sky visibility, making them less effective for indoor use. Recognizing this limitation, researchers have redirected their focus towards alternative wireless infrastructures for localization in indoor spaces, including Wi-Fi, Bluetooth, and Zigbee [3]. Among these, Wi-Fi-based localization systems that use received signal strength (RSS) have gained traction due to the widespread availability of Wi-Fi and the ability of modern embedded and IoT devices to capture Wi-Fi RSS [3].

Wi-Fi RSS-based approaches leverage data from Wi-Fi access points (APs) to determine the positions of mobile and IoT devices within indoor environments. In this context, two methodologies have emerged: propagation model-based and fingerprinting model-based localization systems. Propagation model-based systems employ geometric models such as trilateration and triangulation to determine mobile device location using Wi-Fi RSS [4]. Unfortunately, these systems can be susceptible to inaccuracies from environmental factors such as signal variability caused by obstacles, human interference, signal attenuation, multi-path fading, and shadowing [4]. Fingerprinting model-based systems do not rely on geometric models but instead create a database of Wi-Fi signal patterns ("fingerprints") collected throughout the indoor space. Fingerprinting models are more robust to fading/shadowing/attenuation effects and have demonstrated higher accuracies than propagation model-based methods [3].

Fingerprinting systems consist of two phases: an offline phase for initially constructing the fingerprint database and an online phase for real-time localization. In the online phase, signal patterns may change unpredictably due to factors such as noise in the environment and device variations (devices capturing dissimilar fingerprints at the same location) [5]. These factors can lead to inaccuracies in location estimations.

To address these limitations, machine learning (ML) techniques can be used to learn from data patterns in the offline phase and adjust predictions in the online phase [6]. By combining the strengths of fingerprinting models with ML, researchers have made significant strides in improving the accuracy and reliability of indoor localization solutions [7].

However, ML-based systems are vulnerable to attacks from adversaries who can exploit vulnerabilities in ML models and subtly alter input data to launch adversarial attacks [8]. In Fig. 1, we illustrate the impact of a well-known FGSM adversarial attack [27] on three ML-based indoor localization solutions that use K-Nearest Neighbors (KNN) [13], Gaussian Process Classifier (GPC) [14], and Deep Neural Networks (DNN) [15]. The high loss in accuracy from such attacks can lead to dire consequences, as ML-based indoor localization solutions are often relied upon for decision-making [8][9].

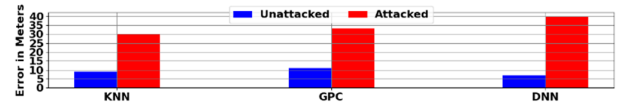

Fig. 1: Accuracy reduction (increase in localization error) in three ML-based indoor localization solutions [13]-[15] due to adversarial attacks

ML-based RSS fingerprinting indoor localization solutions are also susceptible to other security threats such as AP spoofing, AP signal manipulation, and malicious APs [11]. Mitigating the effect of unpredictable RSS fluctuations due to attacks, as well as environmental and device variations, without compromising localization accuracy remains a formidable challenge. Very few prior works address security and robustness in ML-based indoor localization solutions. Even fewer address challenges from adversarial attacks.

In this paper, we present *CALLOC*, a novel framework designed to address the challenges posed by environmental noise, device variations, and the often-overlooked aspect of adversarial security during indoor localization. Our novel contributions as part of the *CALLOC* framework are:

- We develop a novel curriculum learning technique, akin to a knowledgeable teacher guiding a student through progressively complex subjects, to systematically enhance the ML model's resilience to RSS fluctuations.
- We propose a lightweight domain-specific ML model based on scaled dot product attention neural networks for location prediction on resource constrained devices.
- We perform extensive analysis on real-world data to evaluate *CALLOC* under different adversarial attacks,

mobile devices, and building floorplans, and contrast our performance against the state-of-the-art approaches.

## II. RELATED WORK

Various classical ML algorithms such as Naïve Bayes [12], KNN [13], and GPC [14] have been investigated to tackle challenges arising from RSS fingerprint variations due to environmental noise from changes over time in density of people, movement of equipment, etc. However, these approaches do not effectively address issues related to device heterogeneity and adversarial attacks. Device heterogeneity refers to situations where two devices capturing Wi-Fi RSS fingerprints at the same location and time exhibit dissimilar fingerprint patterns, due to disparities in hardware (Wi-Fi chipset) and firmware (noise filtering stack).

More sophisticated ML algorithms based on deep neural networks (DNNs) [15] and convolutional neural networks (CNNs) [16] have emerged in recent years to tackle the complex interplay of environmental noise factors and device heterogeneity. Examples include ANVIL [17][18] which incorporates a multi-headed attention layer with DNNs, SANGRIA [19] which integrates a domain-specific stacked autoencoder and a categorical gradient-boosted tree classifier, WiDeep [14] which combines a de-noising autoencoder with GPC, and the VITAL framework in [20] which uses vision transformers. These advanced ML approaches are shown to outperform their classical ML counterparts when mitigating environmental factors and device heterogeneity. Nevertheless, these works do not consider the crucial challenges associated with adversarial attacks, which are intended to mislead the ML model and are more challenging to address.

Few efforts have addressed attacks on indoor localization solutions. The frameworks in [22], [23] and [25] address the challenges associated with random AP attacks and data privacy but fail to address challenges from adversarial attacks or environmental and device variations. Adversarial attacks, which involve introducing small perturbations into input data, have been explored in various domains, including computer vision, natural language processing, and edge computing, but not as much for indoor localization [8].

To the best of our knowledge, the only prior works that focus on adversarial attacks during indoor localization are [24] and [26]. Advloc [24] leverages DNNs and incorporates a few adversarial samples into the offline training phase to bolster the ML model's resilience against adversarial attacks in the online phase. The work in [26] also employs DNNs but considers a solution with channel state information (CSI) data instead of RSS, rendering it out of scope for our study, because most mobile and IoT devices cannot capture Wi-Fi CSI data. These approaches are also entirely based on simulation, and thus exhibit limited resilience against real-world adversarial attacks. They also do not consider real-world challenges posed by environmental variations and device heterogeneity, which further reduce indoor localization accuracy.

Our proposed framework in this paper overcomes the limitations of the state-of-the-art. We devise a more robust solution for indoor localization, with curriculum learning, which introduces a natural learning order, starting with simpler examples or lessons and progressing toward more intricate ones, thereby enhancing the model's overall learning efficiency progressively. Further, we design a domain-specific attention neural network modified to be lightweight for mobile and IoT device deployment. Our proposed solution is designed to overcome real-world challenges from adversarial attacks and variations across environments and devices.

## III. ADVERSARIAL ATTACK FORMULATION

Adversarial attacks in indoor localization systems encompass the manipulation or spoofing of wireless signals, typically transmitted by Wi-Fi APs. These attacks can originate from three primary points: the transmitter side (i.e., APs), the receiver side (i.e., mobile devices), or the channel side. In the context of Wi-Fi-based indoor localization, Wi-Fi RSS signals, measured in decibels referenced to one milliwatt (dBm) and typically ranging from -100 dBm (weak signal) to 0 dBm (strong signal), are very susceptible to manipulation, particularly when considering attacks from the channel side, (see example in Fig. 2). Channel side attacks offer a distinct advantage due to the open nature of wireless channels, rendering them a preferred choice for adversaries. In contrast, injecting perturbations through the transmitter side (APs) proves more challenging, primarily due to stringent security measures. Meanwhile, attacks on the receiver side necessitate unauthorized access to personal devices, adding an extra layer of complexity. In our considered attack scenario in this work, we concentrate on channel side attacks within a white-box context. Here, we assume that the attacker possesses comprehensive information about building floorplans, Wi-Fi AP locations, and has access to the ML model's training parameters. This white-box approach is motivated by the attacker's capacity to craft subtle perturbations designed to deceive the ML model with minimal alterations.

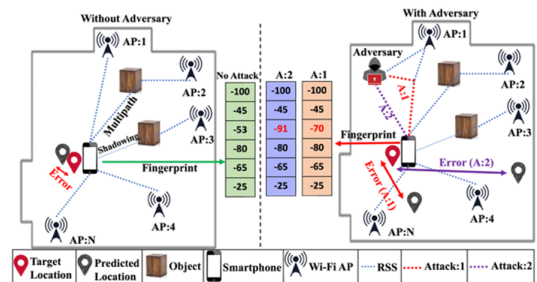

Fig. 2: Illustration of indoor localization system with and without adversarial attacks – A:1 (weak attack), A:2 (strong attack)

### A. Adversarial Attack Mechanism

In this section, we describe the adversarial attack mechanisms we consider in this work that are capable of compromising the robustness of indoor localization systems. Our analysis centers on man-in-the-middle (MITM) attacks from the channel side. MITM attacks are a well-known class of adversarial techniques wherein an adversary positions themselves between communication flows to manipulate data exchanges. In the context of indoor localization, an adversary can insert themselves between a mobile device and the targeted Wi-Fi AP used for localization, as shown in Fig. 2. By introducing perturbations to the signals exchanged between the device and APs, the attacker can mislead the ML model (deployed on the mobile device) to provide incorrect location estimates. MITM attacks are challenging to detect, especially in large indoor environments.

To the best of our knowledge, this is the first exploration of MITM-based attacks in indoor localization. MITM attacks offer a distinct advantage in the context of a white box scenario, empowering adversaries to craft potent and precisely targeted channel-side attacks. We explore two well-recognized variants: signal manipulation and spoofing attacks:

- *Signal manipulation attacks:* These attacks involve altering Wi-Fi signals to provide misleading information to the indoor localization system. They can distort the

RSS data from a Wi-Fi AP, leading to incorrect location estimates. Adversaries may use signal manipulation to misguide the ML model by injecting perturbations into the RSS data, as shown in Fig. 2 (A:1).

- *Signal spoofing attacks:* These attacks involve the creation of counterfeit wireless signals designed to closely mimic legitimate APs. In these attacks, adversaries replicate essential characteristics of a target Wi-Fi AP, such as its MAC address and operating channel. By effectively emulating a genuine AP, the spoofing adversary generates their own RSS data, which outwardly resembles that of the legitimate AP. However, the crucial distinction lies in the carefully crafted perturbations introduced into these counterfeit signals. These perturbations are strategically incorporated to mislead the ML model, resulting in inaccurate location estimates, as shown in Fig. 2 (A:2). This method contrasts with signal manipulation attacks, where genuine RSS data is directly tampered with, whereas signal spoofing attacks involve the fabrication of signals that closely resemble legitimate ones.

*B. White-Box Adversarial Attack Formulation*

In the context of indoor localization security, as discussed above, the main strategy employed in various attack scenarios involves altering RSS data with carefully planned modifications to deceive the ML model. To recreate these scenarios in a real-world white-box environment, we employ three popular methods for crafting white-box adversarial attacks. These methods include FGSM [27], PGD [28], and MIM [29]. Each of these techniques serves a unique purpose in aiding our understanding of adversarial attacks on indoor localization systems. These techniques are described below:

- *Fast gradient sign method (FGSM):* FGSM is a one-step, non-iterative attack method that calculates the adversarial perturbation based on the gradient of the ML model loss function. It then applies this perturbation directly to the input data. It is employed in indoor localization to perturb the RSS data from a targeted Wi-Fi AP. The attack focuses on manipulating RSS data at a single step by adding small but deliberate changes. The FGSM attack can be mathematically represented as follows:

$$X_{Adv} = X + \epsilon * sign(\nabla J(X,Y)) \quad (1)$$

where $X_{Adv}$ is the adversarial example (perturbed RSS data), $X$ is the original RSS data, $\epsilon$ is the magnitude of the perturbation, $\nabla J(X,Y)$ are the gradients of the loss function, and $Y$ is the true label (actual location).

- *Projected gradient descent (PGD):* PGD is an iterative adversarial attack method that refines the perturbation in multiple steps. In each step, it calculates the gradients of the ML model loss function with respect to the input data and updates the perturbation. This iterative process aims to find a stronger perturbation for deceiving the model. Unlike FGSM, which operates in a single step, PGD iteratively enhances the perturbation. The PGD attack can be mathematically represented as follows:

$$X_{Adv} = clip\{X + \alpha * (\nabla J(X,Y), \epsilon)\} \quad (2)$$

where $X$ (original RSS data) is adjusted iteratively. The perturbations are controlled by $\alpha$ (alpha) and capped (*clip function*) at a perturbation magnitude $\epsilon$.

- *Momentum iterative method (MIM)*: MIM is another iterative adversarial attack approach designed to refine perturbations across multiple steps. In each iteration, it computes the gradients of the ML model loss function relative to the input data and updates the perturbation, gradually strengthening it to deceive the model effectively. MIM distinguishes itself from PGD by introducing momentum into the optimization process, which allows it to converge more efficiently and find effective perturbations, making it particularly potent for crafting adversarial examples. The MIM attack can be mathematically represented by equation (2) from PGD.

*C. Adversarial Data Generation*

To evaluate the indoor localization systems' resilience against MITM attacks and their variants, we harness FGSM, PGD, and MIM to craft adversarial data. These techniques allow us to generate MITM-like attacks and systematically assess the robustness of our system. One crucial parameter in crafting adversarial examples is $\epsilon$, representing the attack strength. Varying $\epsilon$ from 0.1 to 0.5 allows us to examine the impact of different attack strengths on the system. As we increase $\epsilon$, the attack strength intensifies, causing perturbations in RSS data to become more pronounced, as shown in Fig. 2 (A:2 has higher $\epsilon$ than A:1). By encompassing this range of $\epsilon$ values, we are able to explore the system's response to a spectrum of adversarial intensities.

Additionally, we consider the parameter ø, which denotes the number of targeted Wi-Fi APs chosen for attack. The example in Fig. 2 shows the case of ø = 1. This selection of specific APs is essential as it mirrors the attacker's choice in the real world. Crafting adversarial data for this subset of APs allows us to evaluate the model's robustness in scenarios where certain APs are compromised. By assessing the system's performance against MITM attacks with varying $\epsilon$ and ø, we can gain insights into a frameworks capacity to withstand adversarial threats across different conditions.

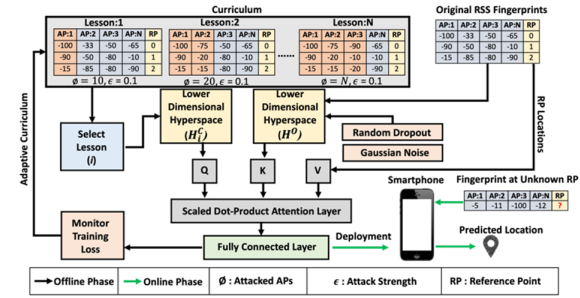

Fig. 3: Overview of the *CALLOC* framework: offline and online phases

## IV. *CALLOC* FRAMEWORK

The *CALLOC* framework consists of two phases: an offline and an online phase, as shown in Fig. 3.

In the offline phase, we begin by collecting RSS fingerprints from various reference points (RPs) in a building using a single device (see section V.A). Initially, this data collection occurs without any adversarial interference (ø = 0, $\epsilon = 0$). To enhance model robustness against adversarial threats, we employ a novel curriculum that systematically introduces attacks on different APs (ø) (see section IV.A). This curriculum has a dual purpose: it guides the model's learning trajectory and progressively challenges it with escalating adversarial scenarios by targeting different subsets of APs (increasing ø).

Once the lesson is selected from the curriculum, we utilize domain-specific neural networks (see section IV.B) to transform both the lesson and the original attack-free data into lower-dimensional hyperspaces: $H_i^C$ (curriculum data) and $H^O$ (original data), as shown in Fig. 3, where $i$ is the lesson number. $H_i^C$ captures essential curriculum features, while $H^O$ captures original data features along with random dropouts and Gaussian noise, to simulate real-world environmental and device variations. This augmentation enhances the model's adaptability to practical scenarios.

These hyperspaces, along with RP locations, serve as inputs for the attention layer within the model, requiring $Q$ (query), $K$ (key), and $V$ (value) inputs (see section IV.C). $H_i^C$ is assigned to $Q$, $H^O$ to $K$, and RP locations to $V$. The attention mechanism enables the model to focus on relevant information within the hyperspaces and RP locations, allowing it to capture crucial patterns that contributes towards accurate indoor localization predictions. The attention output is then passed to a fully connected layer responsible for predicting RP locations for unknown fingerprints collected during the online phase, as shown in Fig. 3. Throughout the training process, we closely monitor the loss function of the final fully connected layer (see section IV.D). This monitoring continues until the final lesson is completed.

In summary, *CALLOC*'s approach involves a novel curriculum-based learning and domain-specific hyperspace transformations with attention, to simultaneously build resilience to adversarial threats, environmental variations, and device heterogeneity, while being lightweight for deployment on mobile and IoT devices. The following subsections describe the framework in more detail.

### A. Curriculum Selection

Our curriculum comprises of 10 distinct lessons. The curriculum initiation begins with a baseline lesson featuring 0% attacked APs (ø = 0) and 100% original data. Subsequent lessons contain higher ø and lower number of original data. For example, the second lesson contains ø = 10 (10% attacked APs) with ϵ = 0.1. This progression culminates in the toughest scenario at lesson 10, with ø = 100 and ϵ = 0.1. By maintaining a fixed and small ϵ value during training, we simulate subtle perturbations similar to those an adversary might introduce in the online phase. As we transform this perturbed data into a lower-dimensional hyperspace (see section IV.B), we capture nuanced changes in essential features within the curriculum data (influenced by ϵ) compared to the original data. This guides the model to learn how perturbations can affect the RSS inputs. Consequently, *CALLOC* becomes resilient against similar perturbation patterns, even when the attack magnitude (ϵ) varies, ultimately enhancing its robustness during testing.

Adversarial data is generated using the FGSM technique, ensuring that the model learns to defend against adversarial attacks. Note that our approach does not require exposure to the many other types of adversarial data (e.g., from PGD and MIM) during training, while still showing robustness to those attacks in the online phase (see Section V). The selection of lessons follows a systematic approach, commencing with lesson 1, where the model learns to handle original data effectively. After completing each training phase, the model proceeds to the next lesson in a progressive manner until it has undergone training across all the lessons.

### B. Lower-Dimensional Hyperspace

Once a lesson is selected for training, the next step involves mapping the lesson data into a lower-dimensional hyperspace, a crucial step in our training approach. This mapping aims to retain essential information while reducing the dimensionality of the input data, making it more computationally efficient and reducing the risk of overfitting. We employ two specialized embedding neural networks for this purpose: one dedicated to the curriculum data and another to the original data. These networks are designed with a reduced number of neurons compared to the input features. The primary motivation behind this dimensionality reduction is twofold. First, it enhances the model's ability to generalize effectively, by focusing on essential features and patterns in the data. Second, reducing dimensionality serves as a means of noise reduction. Noise in the data can originate from environmental factors or device variations. Additionally, these hyperspaces offer computational efficiency, a crucial consideration, especially for mobile and IoT device.

The embedding network for the original data incorporates dropout and Gaussian noise layers. Dropout randomly removes some neuron outputs during training, preventing the model from relying too heavily on certain input features, thus preventing overfitting. The Gaussian noise layers add variability to the training process, further enhancing the model's robustness across different conditions.

### C. Scaled Dot-Product Attention Neural Network

The hyperspaces generated from the curriculum data and the original data represent lower-dimensional representations of the input data, respectively. To measure similarities between these hyperspaces, we employ a scaled dot-product attention neural network. This attention mechanism takes three inputs: $Q$ (query), $K$ (key), and $V$ (value). By computing similarities between elements in $Q$ ($H_i^C$) and $K$ ($H^O$), the attention layer enables the network to prioritize relevant features in the hyperspaces and their corresponding RP locations. This mathematical process can be expressed as:

$$Attention(Q, K, V) = Softmax\left(Q\,K^T / \sqrt{d_k}\right) V \quad (3)$$

In this equation, $Q$ represents $H_i^C$, $K$ represents $H^O$, and $V$ signifies RP locations. The dot product between $Q$ and $K$, followed by scaling ($\sqrt{d_k}$), generates attention scores. These scores, after applying the *Softmax* function, become the attention weights that capture the model's focus on relevant information. Higher attention weights suggest that the model is giving more importance to certain elements in the hyperspaces. The attention output, consisting of weighted information, is then channeled to a fully connected layer. This layer is responsible for classifying and predicting RP locations for unknown fingerprints in the online phase. By utilizing the attention-derived insights, the fully connected layer can make precise predictions in real-world scenarios.

### D. Adaptive Curriculum

Throughout the training process, we carefully monitor the loss function of the final fully connected layer. This monitoring process is vital as any observed increase in loss during training can be an early indicator of a potential divergence in the training process. Divergence may indicate that the model is struggling to adapt to specific data patterns, which can be influenced by ø in the lessons. In response to this, *CALLOC* employs an early stopping mechanism to safeguard optimal performance. When such an increase in loss is detected, the model reverts to its best-performing weights, ensuring that it maintains peak performance.

Moreover, *CALLOC*'s curriculum is not static; it is a responsive component that fine-tunes the model's training approach. After reverting to the best weights, the curriculum is carefully adjusted by reducing ø by steps of two in the lesson data. This adjustment allows for the management of data complexity and further improves the model's adaptability. The curriculum adapts alongside the model, ensuring that the training process is tailored to the evolving needs of the model. Once the training process successfully reduces loss, indicating improved adaptation, the model advances to the next lesson. This dynamic procedure continues iteratively until the final lesson is completed, resulting in a model well-prepared to handle uncertainties in real-world scenarios.

## V. EXPERIMENTAL RESULTS

### A. Experimental setup

In our experiments, we evaluate *CALLOC*'s performance in real-world scenarios across various adversarial attacks. We also compare *CALLOC* against state-of-the-art frameworks, including AdvLoc [24], SANGRIA [19], ANVIL [17], and WiDeep [15]. *CALLOC* exclusively trains on adversarial samples generated via FGSM, maintaining a small constant $\epsilon$ value of 0.1 throughout the curriculum while varying ø. Furthermore, we conduct an extensive analysis of *CALLOC*'s resilience across diverse mobile devices and building floorplans. The details of the heterogeneous mobile devices selected for evaluation are presented in Table I.

TABLE I: SMARTPHONES DETAILS

| Manufacturer | Model | Acronym |
|---|---|---|
| BLU | Vivo 8 | BLU |
| HTC | U11 | HTC |
| Samsung | Galaxy S7 | S7 |
| LG | V20 | LG |
| Motorola | Z2 | MOTO |
| Oneplus | 3 | OP3 |

The building floorplans considered in our experiments encompass variations in path length, visible APs, and environmental noise characteristics (see Table II). Data collection involves the devices selected, with an allocation of 5 fingerprints per RP per building for training and reserving 1 fingerprint per RP per device per building for testing purposes. The OP3 device serves as our designated device to capture training data. The RPs maintain a physical granularity of 1 meter. *We plan to open-source the dataset from this study, to benefit the indoor localization community* [10].

TABLE II. BUILDING FLOORPLAN DETAILS

| Building Number | Visible APs | Path Length | Characteristics |
|---|---|---|---|
| Building 1 | 156 | 64 meters | Wood and Concrete |
| Building 2 | 125 | 62 meters | Heavy Metallic Equipments |
| Building 3 | 78 | 88 meters | Wood, Concrete, Metal |
| Building 4 | 112 | 68 meters | Wood, Concrete, Metal |
| Building 5 | 218 | 60 meters | Wide Spaces, Wood, Metal |

*CALLOC* utilizes domain-specific embedding neural networks, each composed of 128 neurons, utilizing mean square error as the loss function for both hyperspaces. In the $H^O$ network, we empirically determine and set dropout rate to 0.2 and Gaussian noise to 0.32. The final fully connected layer is designed to classify the RP classes, resulting in a total of 65,239 trainable parameters. These parameters are distributed as follows: 42,496 trainable parameters in both the embedding layers, 18,961 in the attention layer, and 3,782 in the final fully connected layer, achieving a compact model size of 254.84 kB (kilobytes), for efficient deployment.

### B. CALLOC Evaluations: Devices, Floorplans and Attacks

In this section, we assess *CALLOC*'s performance across smartphones and buildings under different adversarial attack methods. Fig. 4 presents heatmaps corresponding to changes in indoor localization accuracy under FGSM, PGD, and MIM attacks. The heatmaps display the mean localization error in meters. The experiments are conducted over varying $\epsilon$ (ranging from 0.1 to 0.5) and ø (ranging from 10 to 100) values, to represent diverse real-world adversarial attack scenarios. The tests across multiple buildings are meant to capture a variety of environmental noise scenarios. Lastly, the model is trained on the OP3 device and tested on all devices, to capture scenarios of noise due to device heterogeneity.

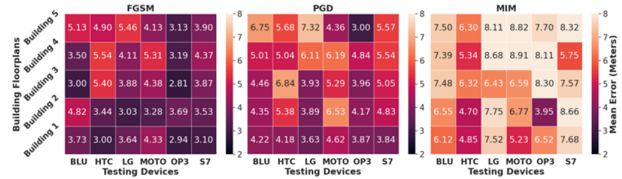

Fig. 4: Localization errors across different devices, buildings, and attacks

In general, as expected, localization accuracy for *CALLOC* reduces in the presence of attacks, environmental variations, and device heterogeneity. However, *CALLOC* is able to limit the accuracy degradation in the presence of variations (as will also become apparent in later results that contrast it with the performance of other frameworks). The heatmaps show that *CALLOC* demonstrates resilience to device heterogeneity, as it maintains consistent performance regardless of the testing devices (i.e., low errors across a row in the heatmaps). Specific devices do exhibit higher variations in error rates e.g., FGSM-Building1-OP3 and FGSM-Building1-MOTO. This behavior can arise due to extreme heterogeneity in wireless chipsets and software stacks across devices, that can cause the captured RSS data from these devices to vary significantly. Additionally, certain building floorplans may exhibit higher errors compared to others, e.g., Building 1 and Building 5. This behavior is due to different salient features in the building floorplans contributing to added noises (e.g., greater dynamic density of people, movement of equipment). Across the three attacks, FGSM has the least error, while stronger iterative attacks such as PGD and MIM result in slightly higher errors compared to FGSM. Nonetheless, *CALLOC*, equipped with its specialized curriculum and hyperspace-attention model, effectively reduces the impact of device heterogeneity and environmental noise, as well as unpredictable adversarial attacks, by limiting the increase in indoor localization error.

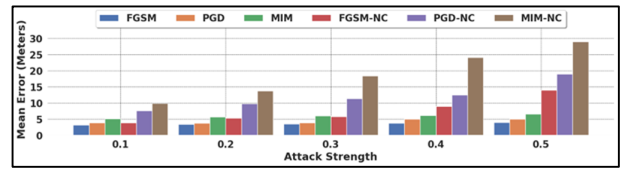

Fig. 5: Impact of curriculum learning on different attacks and $\epsilon$

## C. Evaluating Impact of Curriculum Learning with ϵ

Next, we assess the impact of the proposed curriculum learning technique across varying attacks and ϵ values. Fig. 5 presents a plot depicting the responses of different attacks when subjected to different ϵ values. Each bar represents the mean response across all testing devices, building floorplans, and ø values ranging from 10 to 100. We also compare these results with the performance of the *CALLOC* framework when curriculum learning is not applied, denoted as 'NC' (No Curriculum). The trends from the results reveal that the incorporation of curriculum learning bolsters *CALLOC*'s resilience against varying ϵ values and diverse attack methods, outperforming the 'NC' approach. Notably, in the absence of curriculum learning, the model exhibits pronounced susceptibility to different attacks, particularly in scenarios with higher ϵ values. These findings highlight the efficacy of curriculum learning to withstand adversarial attacks and variations in ϵ values.

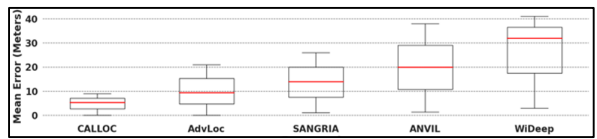
Fig. 6: Comparing CALLOC against state-of-the-art frameworks.

## D. Comparing CALLOC against state-of-the-art

We evaluate *CALLOC's* performance against state-of-the-art frameworks across different devices, buildings, ϵ (0.1 to 0.5), and ø (1 to 100), as shown in Fig. 6. It can be observed that *CALLOC* consistently outperforms its competitors, showing the lowest maximum (worst-case), and mean errors. Notably, *CALLOC* surpasses AdvLoc by 1.77× (mean) and 2.35× (worst-case), due to its novel curriculum learning strategy and domain-specific hyperspace-attention model. It also outperforms SANGRIA, which excels in augmentation to noise but lags in adversarial robustness, by 2.64× (mean) and 2.92× (worst-case). *CALLOC* outperforms ANVIL, which provides exceptional heterogeneity and noise resilience due to its strategic multi-headed attention network, but lags in mitigating adversarial attacks, by 3.77× (mean) and 4.26× (worst-case). *CALLOC* outperforms WiDeep by a remarkable 6.03× (mean) and 4.6× (worst-case). This is due to the incorporation of GPC in WiDeep, which is extremely sensitive to noise. These results highlight *CALLOC's* exceptional performance in real-world scenarios.

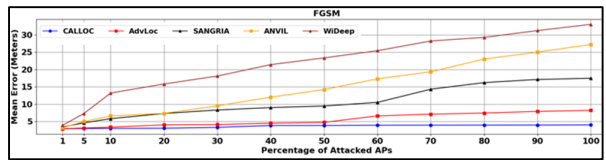
Fig. 7: Effect of *v*arying attacked APs (Ø) on *l*ocalization *e*rror

## E. Evaluating impact of varying ø

We further evaluate *CALLOC's* response to variations in the number of attacked APs (ø) using different attack methods, covering ø values ranging from 1 to 100, and compare the results with state-of-the-art frameworks. Fig. 7 illustrates *CALLOC's* response for different ø values alongside the performance of other frameworks. It is evident that even small increments in ø can significantly reduce localization accuracy. *CALLOC*, in response to the FGSM attack, exhibits relatively stable errors as ø increases, unlike other frameworks. AdvLoc, another framework that also incorporates a subset of FGSM samples into its training, was surpassed by *CALLOC* across FGSM data and more potent attacks like PGD and MIM. AdvLoc shows a slightly high error than *CALLOC*, with the error increasing starting at ø = 60. ANVIL, SANGRIA, and WiDeep show higher errors for both low and high values of ø. These trends hold for PGD and MIM attacks as well (result plots omitted for brevity). In summary, *CALLOC's* resilience to adversarial attacks set it apart from other frameworks from prior work.

## VI. CONCLUSION

Our proposed *CALLOC* framework showcases resilience against adversarial attacks, variations across devices, and noise across indoor environments. It surpasses achievable accuracy over state-of-the-art localization frameworks by a substantial margin, showing improvements of up to 6.03× in mean errors and 4.6× in worst-case errors. Through rigorous evaluation, we observed that *CALLOC* consistently maintains lower localization errors across devices, buildings, and adversarial attack vectors. The implementation of curriculum learning significantly enhances *CALLOC's* robustness, while being lightweight for mobile and IoT device deployment.

## ACKNOWLEDGEMENTS

This research is supported in part by the National Science Foundation grant CNS-2132385.